\documentclass{article}

\usepackage{PRIMEarxiv}

\usepackage[utf8]{inputenc} 
\usepackage[T1]{fontenc}    
\usepackage{hyperref}       
\usepackage{url}            
\usepackage{booktabs}       
\usepackage{amsfonts}       
\usepackage{nicefrac}       
\usepackage{microtype}      
\usepackage{lipsum}
\usepackage{graphicx}
\graphicspath{{media/}}     
\usepackage[caption=false,font=footnotesize]{subfig}
\usepackage{comment}
\usepackage{tikz}
\usepackage{hyperref}
\hypersetup{
    colorlinks=true,
    linkcolor=blue,
    filecolor=magenta,      
    urlcolor=cyan,
    citecolor=blue,
    }
\usepackage{url}
\usepackage{xcolor}
\usepackage{wrapfig}
\usepackage[caption=false,font=footnotesize]{subfig}
\makeatletter
\newcommand*{\rom}[1]{\expandafter\@slowromancap\romannumeral #1@}
\makeatother
\usepackage{amsmath}
\usepackage{algpseudocode}
\usepackage{amsthm}
\usepackage{dsfont}
\usepackage{algorithm}
\usepackage{thmtools}
\usepackage{amssymb}

\def\mbb{\mathbf{b}}

\def\mbx{\mathbf{x}}
\def\mby{\mathbf{y}}

\def\mbI{\mathbf{I}}

\def\mbW{\mathbf{W}}
\def\mbX{\mathbf{X}}
\def\mbY{\mathbf{Y}}

\theoremstyle{definition}

\algnewcommand\algorithmicinput{\textbf{Input:}}
\algnewcommand\Input{\item[\algorithmicinput]}
\algnewcommand\algorithmicoutput{\textbf{Output:}}
\algnewcommand\Output{\item[\algorithmicoutput]}
\algnewcommand\algorithmicinit{\textbf{Initialize:}}
\algnewcommand\Init{\item[\algorithmicinit]}

\definecolor{pleasant_red}{HTML}{FFB6C1}  
\definecolor{golden_color}{HTML}{FFD700}   
\definecolor{spring_green}{HTML}{00FF7F}   
\title{Trust, but Verify:\\ Peeling Low-Bit Transformer Networks for Training Monitoring}

\author{
  Arian Eamaz \\
  Department of Electrical and Computer Engineering \\
  University of Illinois Chicago \\
  Chicago, IL, USA \\
  \texttt{aeamaz2@uic.edu} \\
  \And Farhang Yeganegi \\
  Department of Electrical and Computer Engineering \\
  University of Illinois Chicago \\
  Chicago, IL, USA \\
  \texttt{fyegan2@uic.edu} \\
  \And Mojtaba Soltanalian \\
  Department of Electrical and Computer Engineering \\
  University of Illinois Chicago \\
  Chicago, IL, USA \\
  \texttt{msol@uic.edu} \\
}

\begin{document}
\maketitle

\begin{abstract}
Understanding whether deep neural networks are effectively optimized remains challenging, as training occurs in highly nonconvex landscapes and standard metrics provide limited visibility into layer-wise learning quality. This challenge is particularly acute for transformer-based language models, where training is expensive, models are often reused in frozen form, and poorly optimized layers can silently degrade performance. We propose a layer-wise peeling framework for monitoring training dynamics, in which each transformer layer is locally optimized against intermediate representations of the trained model. By constructing lightweight, layer-specific reference solutions and projecting layers onto multiple intermediate outputs via different permutations, we obtain achievable baselines that enable fine-grained diagnosis of under-optimized layers. Experiments on decoder-only transformer models show that these layer-wise reference bounds can match or even surpass the trained model at various stages of training, exposing inefficiencies that remain hidden in aggregate loss curves. We further demonstrate that this analysis remains effective under binarization and quantized settings, where training dynamics are particularly fragile. Across all numerical results, the proposed bounds consistently separate apparent convergence from effective optimality, highlighting optimization opportunities that are invisible when relying on training loss alone.
\end{abstract}

\section{Introduction}

Deep neural networks have become a central tool in modern machine learning and signal processing due to their ability to construct expressive representations through deep nonlinear compositions \cite{lecun2015deep,goodfellow2016deep}. Despite their empirical success, understanding and verifying the quality of the training process itself remains a fundamental challenge \cite{srivastava2024optimistic}. In practice, optimization is performed in highly nonconvex landscapes, the global optimum is unknown, and standard training curves provide limited insight into whether learning dynamics are genuinely effective or merely progressing toward suboptimal or unstable solutions \cite{choromanska2015loss, dauphin2014identifying}. In particular, models may appear to converge in terms of loss while still failing to exploit their representational capacity \cite{zhang2017understanding, neyshabur2017geometry}. This issue becomes especially critical as models grow deeper, wider, and more expensive to train \cite{hernandez2021scaling,hoffmann2022training}.

The challenge is further amplified in the context of large language models. Modern language models are typically trained once at massive scale and then reused in frozen form for downstream tasks \cite{touvron2023llama, brown2020language}. In such settings, practitioners rely on fine tuning, adapters, or lightweight heads while assuming that the underlying pretrained representations are reliable \cite{hu2022lora,houlsby2019parameter}. However, without principled tools to diagnose training fidelity, it is difficult to justify confidence in either the original training process or subsequent adaptations. Since training trajectories, optimizer states, and intermediate checkpoints are often unavailable, diagnosing optimization failures must be done post hoc, using only the trained model and data \cite{sardana2023beyond, roberts2020much}. Small inefficiencies, poorly conditioned layers, or optimization instabilities can propagate silently through the network, with no clear mechanism to detect them.

A large body of prior work has studied convergence diagnostics and optimization dynamics, primarily from a statistical or solver specific perspective. Existing approaches focus on learning rate schedules, gradient norm analysis, or theoretical convergence guarantees for stochastic optimization methods such as stochastic gradient descent and its variants \cite{bottou2018optimization,goyal2017accurate,park2022generalization, rajput2020closing,pesme2020convergence,allen2019convergence}. While valuable, these methods are often indirect, hyperparameter dependent, and do not provide data aware certificates of training quality. Moreover, they typically treat the network as a monolithic object and offer limited insight into where within a deep architecture training may be underperforming, or how close the observed solution is to a simple and achievable baseline. 

To address this gap, the authors of \cite{yeganegi2025data} introduce the YES training bounds, a data driven framework for certifying neural network training performance in real time. The core idea is to compare the loss achieved by the optimizer against a simple layer wise reference solution constructed via projection based optimization. If the training trajectory fails to surpass this reference, the optimizer is provably underperforming. If it does, one obtains a certificate that meaningful learning has occurred beyond what is achievable by simple constructions. These bounds are solver agnostic, data dependent, and provide transparent feedback throughout training without modifying the objective, gradients, or optimization procedure. In their original formulation, YES bounds were developed for fully connected networks with linear projections, a setting that is analytically tractable and remains a fundamental building block of deep architectures.

In this paper, we extend the YES framework beyond its original scope. While fully connected networks remain important as canonical models for representation learning, modern architectures, particularly transformer based language models, introduce new structural and dynamical complexities that demand careful scrutiny. We generalize the YES framework to more sophisticated models, including transformer based language models, and study how layer-wise training quality evolves in these settings. Our approach adopts a layer by layer peeling strategy, in which each layer is locally optimized to form a reference solution, enabling fine grained inspection of training dynamics across depth. This allows us to identify anomalous layers, diagnose optimization bottlenecks, and assess whether deeper components contribute meaningfully beyond simple baselines. 

Related work analyzes intermediate representations using auxiliary classifiers or probes. The authors of \cite{alain2016understanding} introduce linear classifier probes to measure linear separability across network depth. While informative about representational content, such probes are not designed to assess optimization quality: they introduce additional learned components, depend on solver choices, and do not certify whether training has surpassed a simple or achievable baseline.

Other studies show that internal representations often stabilize early in training even as loss continues to decrease. For example,  \cite{morcos2018insights} reports high representational similarity across layers early in optimization. These results suggest that apparent convergence can mask deeper optimization limitations. Our bounds provide a complementary, optimization-level view by explicitly detecting suboptimal convergence, including cases where training appears stable under quantization.

Beyond architectural complexity, we also investigate training under low resolution and constrained settings, where optimization is known to be fragile. While quantization is not the primary focus of this work, it serves as a stress test that highlights the need for principled and architecture aware monitoring tools. Across both toy examples such as MNIST and language modeling tasks, our results demonstrate that the proposed bounds provide actionable and interpretable certificates of training fidelity. Unlike probing methods or auxiliary classifiers, which assess the semantic content of learned representations, YES bounds directly evaluate the effectiveness of the optimization process itself. \emph{Our primary objective is not to achieve state-of-the-art language modeling performance or to propose a new optimization algorithm, but rather to enable a controlled evaluation of how well a model can perform under limited data and specific training configurations.} Our main contributions in this paper are:

\begin{itemize}
    \item The proposed YES bounds disentangle optimization inefficiency from quantization-induced distortion, enabling practitioners to identify whether performance bottlenecks arise from poor solver dynamics or from low-precision representations.

    \item YES bounds provide layer wise certificates that evolve during training and reveal how individual layers contribute to optimization. For decoder only transformers, we observe two key effects. First, YES constructions with fewer layers can outperform the main model at certain epochs. Second, different permutations of layer-wise projection targets produce noticeably different performance, with some consistently surpassing the direct projection that follows the original training order. This indicates that training dynamics across layers are not uniformly effective. If training aligns with or exceeds the YES bounds, the optimization path is effective within our framework. If it does not, the intermediate solutions from our construction, while not necessarily optimal, may represent more efficient paths than those taken during standard training.


    \item Our results further show that YES solutions can achieve test performance that is competitive with, and in some cases better than, their training performance. Moreover, improvements in the training quality of the main model are consistently reflected in stronger YES-bound solutions on the test set. This correlation suggests that YES bounds may serve as a principled indicator for training progress and a potential stopping criterion that preserves generalization.
\end{itemize}

\section{Training Performance Certification with Alternative Layer-Wise Solutions: Fully Connected Case}
We begin with a toy example, a ReLU network, to clearly provide an intuitive perspective on our main idea. The core idea of our scheme is to compare the solution obtained through training with a baseline that can be computed in a low cost layer wise manner. If the training trajectory fails to converge to at least this baseline, or if it consistently remains above it, this behavior indicates a potential misconfiguration of the solver, for example due to inappropriate parameter choices such as the learning rate.

This challenge is particularly pronounced in coarsely quantized neural networks, where solvers often lack convergence guarantees and are prone to poor local minima.

\subsection{Alternative Solution for the Last Layer}
Let $\mbW_k$ and $\mbb_k$ denote the weight matrix and bias vector for layer $k\in[K]$ of the network, respectively. Suppose the input and output data are represented by matrices $\mbX\in\mathbb{R}^{n\times d}$ and $\mbY\in\mathbb{R}^{m\times d}$, respectively.
We consider the $\ell_2$-squared training loss for such a network employing the ReLU activation function $\Omega(.)$.

We begin with the last layer. Let $\mbY_{k}^{\star}$ denote the output of the penultimate layer obtained from training. For the binary neural network, rather than relying on the solver’s outcome, we solve the last layer directly:
\begin{equation}
\underset{[\mbW_K]_{i,j}\in \{-\lambda,\lambda\}}{\textrm{minimize}}\big\|\mbY - \mbW_K \mbY_K^{\star} \big\|^2_{\mathrm{F}},
\end{equation}
where $\lambda>0$ denotes the quantization scale. For this optimization problem, a proximal linear solution is given by $\widehat{\mbW}_K = \lambda \, \operatorname{sign}\!\left( \mbY \mbY_K^{\star\dagger} \right)$, 
while an alternative is to employ a first-order iterative method:
\begin{equation}
\mbW^{(i+1)}_K = \lambda \operatorname{sign}\left(\mbW^{(i)}_K + \alpha \left( \mbY - \mbW^{(i)}_K \mbY_K^{\star} \right)\mbY^{\star\top}_K\right).
\end{equation}
where $\alpha>0$ is the step size and $i$ indexes the iterations. 
\subsection{Constructing YES Bounds with Intermediate Points}
Building on the concept of the alternative last-layer solution, we extend the same principle to all layers. Initially, rather than depending on intermediate training outputs, we directly project each layer’s input to the final output. We refer to this construction as the \emph{YES-0 bound}. We named this bound YES-0 because it involves zero intermediate projection points in the space. The purpose of YES-0 is to provide a simple baseline: if the training solver is effective, it should at least outperform this direct projection.

Formally, the YES-0 bound is obtained by solving
\begin{equation}
\underset{[\mbW_k]_{i,j}\in \{-\lambda,\lambda\}}{\textrm{minimize}}\big\|\mbY - \Omega\left(\mbW_k \mbY_k\right) \big\|^2_{\mathrm{F}},~k\in[K-1],
\end{equation}
where the last layer with ($k=K$) does not have an activation function $\Omega$. One possible approximated solution to this problem is the proximal linear projection:
\begin{equation}
\begin{aligned}
\widehat{\mbW}_k = \lambda\operatorname{sign}\!\left( \mbY\mbY_k^{\dagger} \right),~
\mbY_{k+1} = \Omega \!\left( \widehat{\mbW}_k \mbY_k \right),
\end{aligned}
\end{equation}
with $\mbY_1=\mbX$ and $\mbY_{K+1}=\widehat{\mbW}_K\mbY_K$. Note that the YES-0 bound is an easily calculated bound that may be used to immediately detect if proper (not necessarily optimal) training has been carried out, in the sense that the network weights have been meaningfully impacted by the training data. The answer (YES or NO) will provide immediate relief as to whether training is being meaningful at all.

While it is fair to say that we hope that by iterative mapping, we get closer and closer to the output of interest, it has also been observed in various machine learning problems that after extensive training, the output of some inner layers become something meaningful to domain experts. To extend the idea of YES-0 bound, we construct YES bounds by leveraging intermediate outputs from training. The key principle is to project the input of each layer not only to the final output $\mbY$, but also to selected useful intermediate points $\{\mbY^\star_j\}$ uncovered by the training process. This approach enables the construction of a systematic, data-driven baselines that evolves
with the training trajectory.

The construction proceeds in the following steps:
\begin{enumerate}
    \item Set $\mbY_1 = \mbX$ as the network input.
    \item Choose an intermediate output $\mbY^\star_j$ from the training trajectory at some layer $j\in[K-1]$.
    \item Solve the alternative objective for $k\in[j-1]$:
    \begin{equation}
    \begin{aligned}
    \widehat{\mbW}_k = \lambda \, \operatorname{sign}\!\left( \mbY^\star_j \mbY_k^{\dagger} \right),\quad
    \mbY_{k+1} = \Omega \!\left( \widehat{\mbW}_k \mbY_k \right),
    \end{aligned}
    \end{equation}
    which projects $\mbY_k$ to $\mbY^\star_j$, ensuring that the bounds are meaningful under one-bit quantization. 
    \item For the remaining layers $k > j-1$, project each input directly to the final output $\mbY$ using
    \begin{equation}
    \begin{aligned}
    \widehat{\mbW}_k = \lambda \, \operatorname{sign}\!\left( \mbY \mbY_k^{\dagger} \right),\quad\mbY_{k+1} = \Omega \!\left( \widehat{\mbW}_k\mbY_k \right),
    \end{aligned} 
    \end{equation}
    with the last layer given by $\mbY_{K+1}=\widehat{\mbW}_K\mbY_K$.
    \item Compute the resulting error $\|\mbY - \mbY_{K+1}\|_{\mathrm{F}}^2$.
    \item Repeat this process for different choices of intermediate outputs $\mbY^\star_j$, and take the minimum error across all such projections.
\end{enumerate}

\subsection{Gradient-Based Refinements}
While the proximal linear updates in YES bounds are often sufficient, they can occasionally yield loose baselines. To tighten these bounds, we incorporate gradient-based refinements:\\
$\bullet$ \textbf{Linear last layer (first order):}
\begin{equation}
\mbW^{(i+1)}_K = \mbW^{(i)}_K + \alpha \left( \mbY - \mbW^{(i)}_K \mbY_K \right) \mbY_K^{\top}.
\end{equation}
$\bullet$ \textbf{Nonlinear ReLU layers (second order):}
\begin{equation}
\begin{aligned}
\mbW^{(i+1)}_k = \mbW^{(i)}_k +  \left( \left( \mbY^{\star}_j - \Omega\left(\mbW^{(i)}_k \mbY_{k-1}\right) \right) \odot \mbI_{\mbW \mbY > 0} \right) \mbY_{k-1}^{\dagger},
\end{aligned}
\end{equation}
where $\mbI_{\mbW \mbY > 0}$ is the indicator matrix.

In our one-bit setting, the construction of YES bounds differs from the last-layer solution in how quantization is applied. For YES bounds, the proximal quantization operator is enforced only after the algorithm has converged (second-order updates) rather than at every iteration. This design deliberately produces a non-optimal baseline, serving as a conservative reference that the training trajectory should surpass.
Applying quantization at every step could cause the solver to prematurely adapt to the discrete constraints, yielding
a misleadingly good baseline. By deferring quantization, we obtain a solution that is not perfect but meaningful for comparison: effective training should outperform it, or at least not perform worse. In contrast, for the last-layer solution, quantization is applied at each iteration, since the last layer directly determines the output, and a weak or vacuous bound
at this stage would be uninformative.



\section{Baseline-Guided Monitoring of Language Model Training}

$\bullet$ \textbf{Note:} Throughout this paper, the terms \emph{GPT-2 style} and \emph{LLaMA style} refer
to specific architectural design choices rather than particular pretrained
weights or datasets. GPT-2–style models use a decoder-only Transformer with
learned absolute positional embeddings added to token embeddings and a standard
feedforward network. In contrast, LLaMA-style models employ rotary positional
embeddings applied within self-attention, RMSNorm with pre-normalization, and a
gated feedforward network (SwiGLU).

In this section, we describe the construction of the YES bounds for language models in a step-by-step manner. At each stage of training, we first train a standard autoregressive language model, such as a LLaMA- or GPT-style transformer. We then use the trained model as a reference to build a corresponding YES solution, which serves as a low-cost, structured approximation of the model’s internal representations. 

The YES construction proceeds layer-wise. We begin by freezing the parameters of the trained language model and extracting the hidden representations produced at each transformer layer on a small cached subset of the training data. These per-layer activations act as targets that define the YES objective. The YES model itself mirrors the depth of the original transformer but is optimized sequentially, one layer at a time, rather than end-to-end.

The token and positional embeddings are taken directly from the trained language model, ensuring that the baseline operates on the same representational space and isolates training quality at the level of transformer blocks and output mappings. Given these embeddings, the first YES layer is optimized to match the corresponding hidden representation of the trained model. Once this layer is fitted, its parameters are frozen, and its output is used as the input for training the next YES layer. This process is repeated iteratively, ensuring that at each step only a single layer is optimized while all preceding layers remain fixed.

After all YES transformer layers have been fitted in this layer-wise fashion, a lightweight output head is trained on top of the final YES representation using the standard language modeling objective. Optionally, a final normalization layer can also be optimized at this stage. Importantly, during this final step, the YES transformer layers themselves are kept fixed, so that the head training reflects the quality of the learned representations rather than further representation learning.

This sequential, layer-wise optimization yields a computationally inexpensive solution that closely tracks the internal behavior of the fully trained model. The resulting YES solution provides a principled baseline for monitoring training progress: it captures the best performance achievable under constrained, local optimization and can be directly compared to the outcome of full end-to-end training.

\subsection{Algorithmic Formulation of YES Construction}

Let \( f_\theta \) denote a trained autoregressive language model with \( L \) transformer layers, token embedding \( E \), positional embedding \( P \), and output head \( \mbW_{\text{out}} \). For an input sequence \( \mbx \in \mathcal{V}^T \), we write
\begin{equation}
h^{(0)}(\mbx) = E(\mbx) + P(\mbx),~
h^{(\ell)}(\mbx) = F^{(\ell)}_\theta\!\left(h^{(\ell-1)}(\mbx)\right), 
\end{equation}
where \(\ell = 1,\dots,L\), and \( h^{(\ell)}(\mbx) \in \mathbb{R}^{T \times d} \) denotes the hidden representation at layer \( \ell \). Here, $F^{(\ell)}_\theta : \mathbb{R}^{T \times d} \to \mathbb{R}^{T \times d}$ represents the mapping implemented by the \(\ell\)-th transformer layer of the trained model, including its multi-head self-attention, feedforward network, residual connections, and layer normalization. That is, \( F^{(\ell)}_\theta \) takes the hidden state from the previous layer \( h^{(\ell-1)}(\mbx) \) as input and produces the updated hidden representation \( h^{(\ell)}(\mbx) \). Note that for the LLaMA-style architecture, positional information is not introduced
via additive positional embeddings but is instead incorporated through rotary
positional embeddings (RoPE) within the self-attention mechanism.

\paragraph{Cached Teacher Representations.}  
We fix a small cache \( \mathcal{C} = \{\mbx_i\}_{i=1}^N \) of training sequences and compute, for each \( \mbx_i \in \mathcal{C} \), the corresponding per-layer representations
\[
\left\{ h^{(\ell)}_\theta(\mbx_i) \right\}_{\ell=1}^L,
\]
which are treated as fixed targets throughout the YES construction.

\paragraph{YES Model Definition.}  
The YES model consists of \( L \) transformer layers \( \{ G^{(\ell)}_\phi \}_{\ell=1}^L \), optional embeddings \( \tilde{E}, \tilde{P} \), and a lightweight output head \( \tilde{\mbW}_{\text{out}} \). Its forward recursion is defined as
\[
\tilde{h}^{(0)}(\mbx) = \tilde{E}(\mbx) + \tilde{P}(\mbx), \qquad
\tilde{h}^{(\ell)}(\mbx) = G^{(\ell)}_\phi\!\left(\tilde{h}^{(\ell-1)}(\mbx)\right).
\]

\paragraph{Layer-wise Optimization.}  
The YES layers are optimized sequentially. For layer \( \ell \), all parameters associated with layers \( 1,\dots,\ell-1 \) are held fixed, and the parameters of \( G^{(\ell)}_\phi \) are obtained by minimizing the regression objective
\[
\min_{\phi_\ell} 
\sum_{\mbx_i \in \mathcal{C}}
\left\|
G^{(\ell)}_{\phi_\ell}\!\left(\tilde{h}^{(\ell-1)}(\mbx_i)\right)
-
h^{(\ell)}_\theta(\mbx_i)
\right\|_2^2.
\]
Once optimized, layer \( \ell \) is frozen, and its output is used as the input for layer \( \ell+1 \). 

\paragraph{Output Head Fitting.}  
After all YES layers have been fitted, we optimize the output head (and optionally a final normalization layer) by minimizing the standard language modeling loss
\[
\min_{\tilde{\mbW}_{\text{out}}}
\sum_{(\mbx,\mby) \in \mathcal{C}}
\mathcal{L}_{\text{CE}}\!\left(
\tilde{\mbW}_{\text{out}} \tilde{h}^{(L)}(\mbx),
\, \mby
\right),
\]
while keeping all YES transformer layers fixed.

\paragraph{Resulting YES Solution.}  
After computing the parameters of the transformer layers using the YES procedure, these parameters are plugged into the original cost function, enabling a comparison with the training trajectory.

\subsection{YES Layer Permutations and Target Selection}
We project each transformer layer onto different intermediate outputs of the trained model to obtain multiple valid reference trajectories, which allows us to test whether the observed training dynamics are aligned with any achievable layer-wise configuration, rather than a single fixed ordering.

In the YES framework, each YES layer \(\tilde{h}^{(\ell)}\) is trained to match a hidden representation of the trained model. Let the trained model have \(L\) layers. We construct a YES model with \(L' \le L\) layers, where \(L'\) denotes the number of layers in the YES model. Choosing \(L' \le L\) allows us to build a computationally cheaper, low-cost approximation of the full model, while still capturing the essential hierarchical transformations. This also enables us to probe how much each layer contributes to the final output and to study potential redundancies.

\paragraph{Direct permutation (canonical YES):}  
The direct permutation corresponds to
\[
\pi_\text{direct}(\ell) = \ell, \quad \ell = 1,\dots,L',
\]
where each YES layer is trained on its corresponding model layer. This resembles the original training itself and preserves the layer-wise transformations.

\paragraph{Alternative permutations:}  
We also consider mappings
\[
\pi : \{1, \dots, L'\} \to \{0, \dots, L\},
\]
allowing each YES layer to target the output of any trained layer. The only constraint is that each YES layer may project only to the output of a deeper transformer layer in the trained network; specifically, the output of layer $\ell$ can be projected only to the output of layer $\ell' \ge \ell+1$. The projection here means the following layer-wise objective:
\[
\min_{\phi_\ell} \sum_{\mbx_i \in \mathcal{C}} 
\Big\|
G^{(\ell)}_{\phi_\ell}\!\big(\tilde{h}^{(\ell-1)}(\mbx_i)\big) - h^{(\pi(\ell))}_\theta(\mbx_i)
\Big\|_2^2.
\]
This lets us analyze the significance of each layer, test redundancies, and probe interactions between layers. 

We note that unlike ReLU networks, transformer-based models do not admit a linear closed-form solution. As a result, it is not possible to project intermediate layers directly onto the final target, and consequently the red (linear) region of the YES cloud is absent in this setting. To reduce computational overhead, we also do not define polynomial degrees for the YES bounds. Instead, users select different output-layer permutations to construct YES$k$ baselines. To avoid non-decreasing behavior, the reported YES curves are obtained by taking the minimum value across all previous epochs at each epoch.

\paragraph{Example YES permutations:}  
\begin{itemize}
    \item \textbf{YES1:} \([3,3,3,4]\) — the first three YES layers target the output of the third trained layer, and the last YES layer targets the fourth layer.  
    \item \textbf{YES2:} \([2,2,3,4]\) — the first two YES layers target the output of the second layer, the third YES layer targets the third layer, and the last layer targets the fourth.  
    \item \textbf{YES3:} \([1,3,3,4]\) — the first YES layer targets its corresponding trained layer output, the second and third YES layers target the third layer, and the fourth YES layer targets the fourth layer.  
    \item \textbf{YES4:} \([1,2,3,4]\) — the canonical direct permutation, each YES layer matches its corresponding layer output. 
\end{itemize}

One could examine all possible permutations of the outputs, but the computational cost quickly becomes prohibitive. Instead, users can select a subset of permutations for evaluation during training. As our results will show, placing greater emphasis on the final layers as well as the earliest layers already provides a strong assessment.

If the layer-wise optimization is effective, the direct permutation setting closely mirrors the original training process at intermediate layers. Differences between the YES solution and the trained model then arise primarily from the optimization of the final output layer, $\mbW_{\text{out}}$, which may yield slightly better or worse performance than the original training outcome. \emph{For a well-trained model, we therefore expect the direct permutation to outperform alternative permutations, as it represents the configuration most consistent with the training baseline.}

\section{Numerical Results}

In this section, we evaluate the proposed baselines for analyzing and weighting deep neural networks across tasks and model scales. Our experiments progress from controlled toy settings to small-scale language models trained from scratch, and finally to large-scale pretrained models fine-tuned under challenging optimization regimes. This staged evaluation allows us to isolate the behavior of the proposed method under increasing data and model complexity. We report results on MNIST using a ReLU binary neural network in Appendix~A. This experiment serves as a toy example to further illustrate the main ideas of the paper. Since our primary focus is on transformer-based architectures, we present these results in the appendix rather than the main text. In our numerical experiments, we consider a range of configurations, including training hyperparameters, architectural choices, and quantization settings, to evaluate our bounds across diverse scenarios.

\subsection{Language Models}

\paragraph{Small-scale language modeling.}
We evaluate a lightweight decoder-only LLaMA-style Transformer on the \textbf{WikiText-2} dataset (\texttt{wikitext-2-raw-v1}), which consists of raw Wikipedia articles without additional preprocessing and contains approximately 2.1M tokens for training and 217K tokens for testing. To enable rapid experimentation and controlled analysis of optimization behavior, we restrict this experiment to a small subset of the corpus, using the first 100,000 token IDs for training and the first 10,000 token IDs for testing. Tokenization is performed using the GPT-2 byte-pair encoding (BPE) tokenizer. The lightweight model uses a batch size of 8, a context length of 32 tokens, an embedding dimension of 64, 2 attention heads, a feed-forward dimension of 128, and 4 Transformer layers with gated MLPs (SwiGLU). Dropout is disabled to isolate optimization dynamics, and training is performed using cross-entropy loss for next-token prediction.

\paragraph{Large-scale fine-tuning: OpenLLaMA-3B.}
For full-scale experiments, we fine-tune a pretrained LLaMA model on the \textbf{WikiText-103} dataset (\texttt{wikitext-103-raw-v1}), which contains raw Wikipedia articles and approximately 103M tokens for training and 217K tokens for validation. We evaluate the \textbf{OpenLLaMA-3B} model \cite{openlm2023openllama}, a decoder-only Transformer with
approximately 3 billion parameters, 26 Transformer layers, a hidden dimension
of 3200, an intermediate (MLP) dimension of 8640, and 32 attention heads.
Fine-tuning is performed using sequences of 512 tokens and a LLaMA-style
SentencePiece tokenizer, with the AdamW optimizer.
 All datasets are obtained using the Hugging Face \texttt{datasets} API, which provides standardized and reproducible access to WikiText. Although WikiText-103 is substantially larger than WikiText-2, it remains small relative to the capacity of OpenLLaMA-3B. This makes it a suitable testbed for evaluating whether, under quantization and fine-tuning, a large pretrained model can follow optimization paths that appear stable in terms of training loss yet remain suboptimal. In particular, even on WikiText-103, fine-tuning a 3B-parameter model under quantization allows us to assess whether our YES bounds can reveal the existence of better optimization trajectories beyond those selected by standard training.

\paragraph{Quantization during fine-tuning.}
To further challenge the optimization process and evaluate robustness, we fine-tune the OpenLLaMA-3B model under different quantization schemes applied during training:
\begin{itemize}
    \item Channel-wise 1-bit quantization
    \item Ternary (1.58-bit) quantization \cite{ma2024era}
\end{itemize}
These experiments assess the ability of the proposed baseline to maintain stable training dynamics and competitive performance under aggressive quantization.
\begin{figure*}[t]
	\centering
	\subfloat[]
		{\includegraphics[width=0.5\columnwidth]{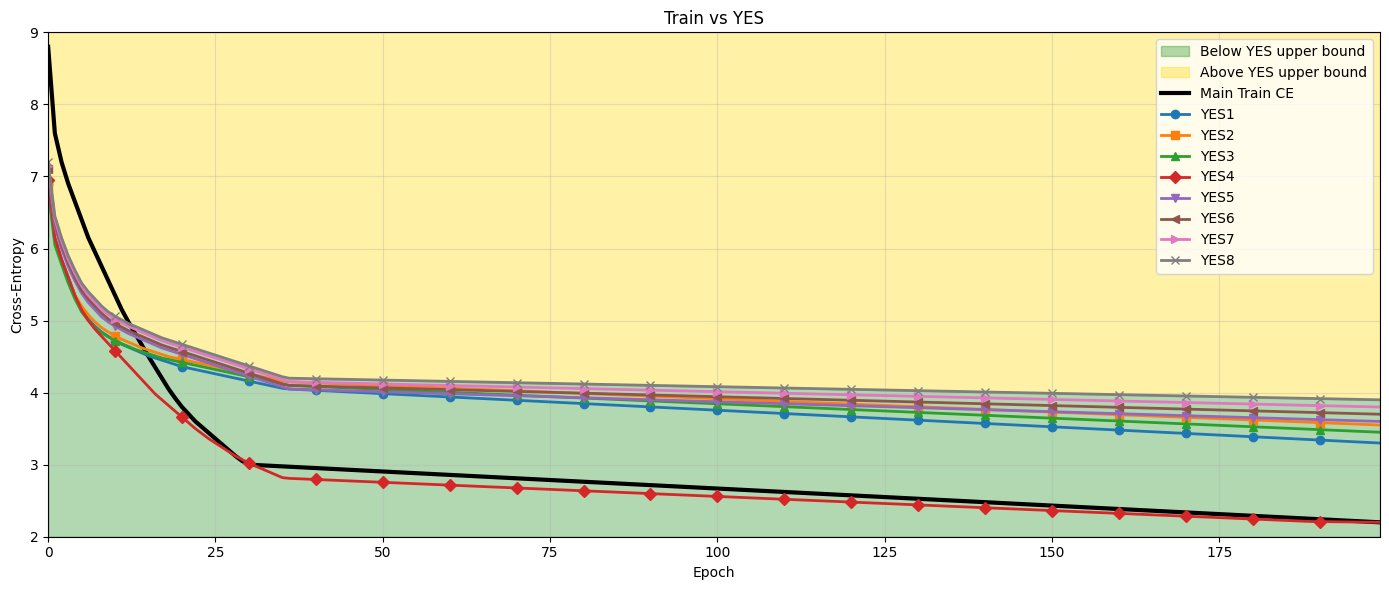}}\hspace{35pt}\quad
    \subfloat[]
		{\includegraphics[width=0.5\columnwidth]{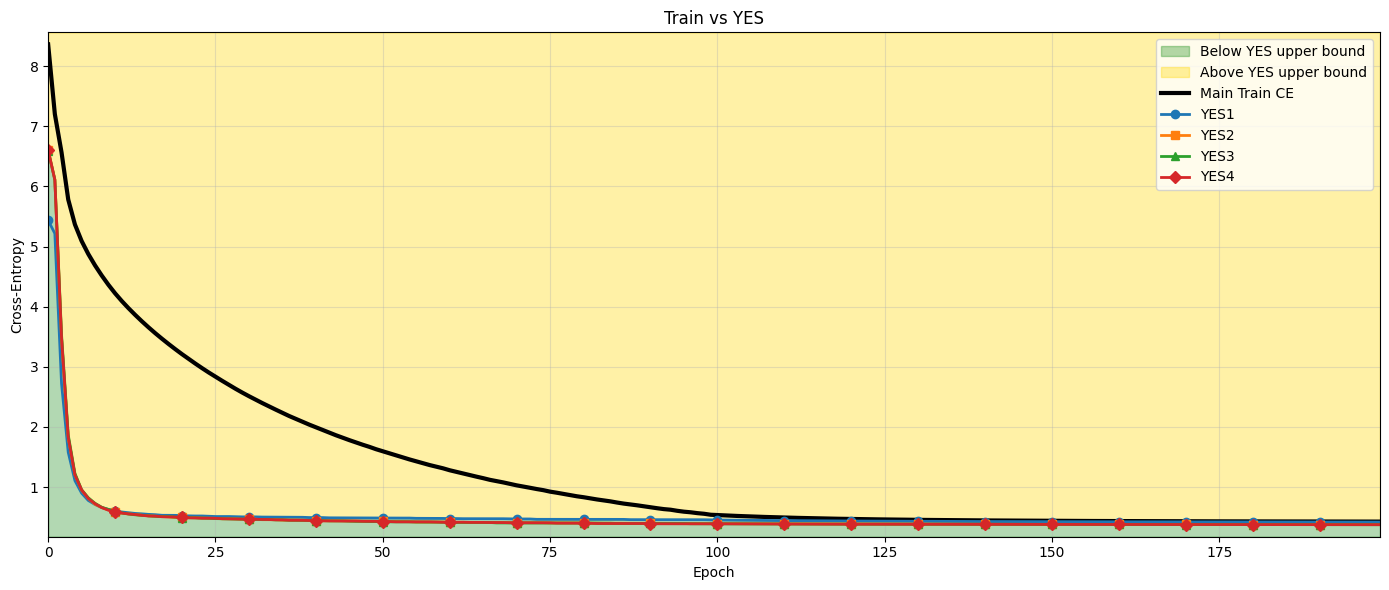}}\\

	\caption{Training trajectories on (a) LLaMA-style (4 layers) and (b) GPT2-style (4 layers) with YES bounds. The plots show how different YES permutations interact with the training process, highlighting the effect of layer importance and learning rate on convergence.
 }
\label{figure_2}
\end{figure*}
\begin{figure*}[t]
	\centering
	\subfloat[]
		{\includegraphics[width=0.45\columnwidth]{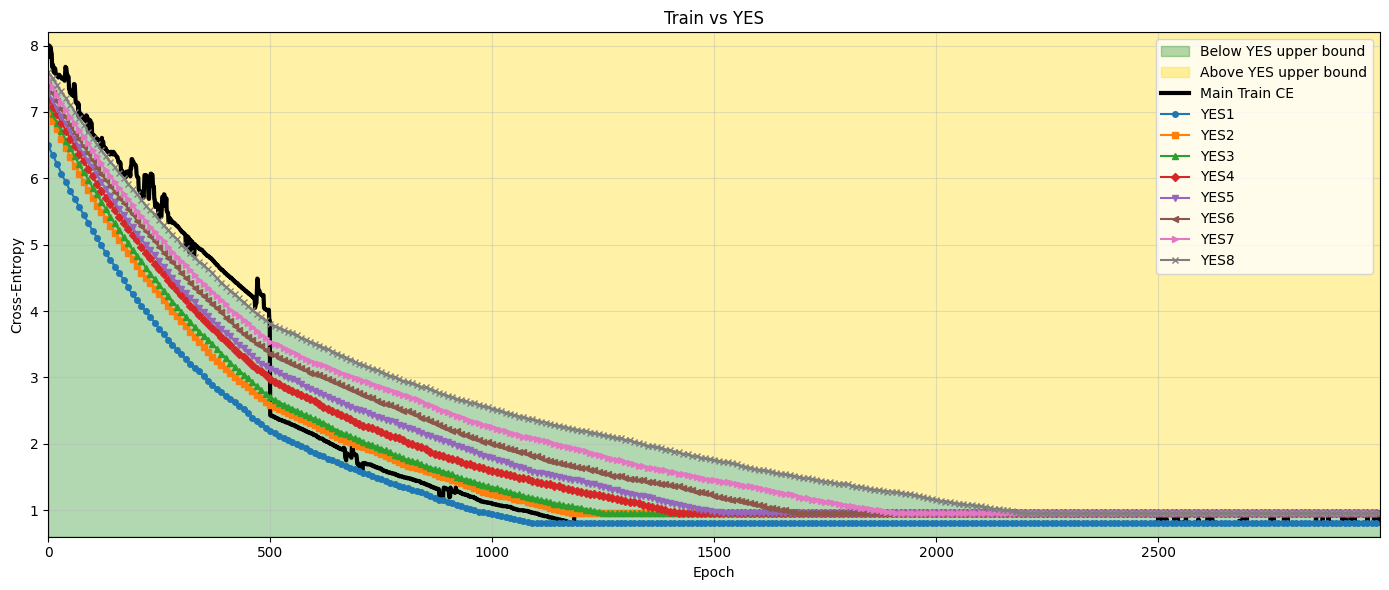}}\hspace{35pt}\quad
    \subfloat[]
		{\includegraphics[width=0.45\columnwidth]{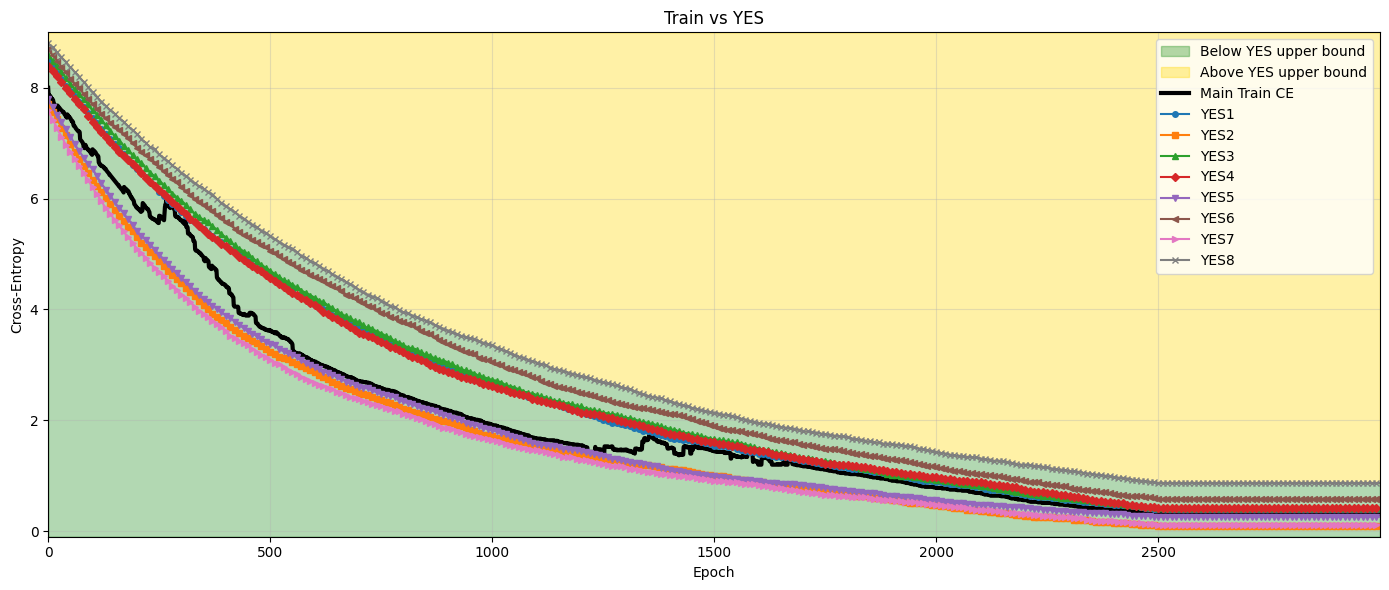}}\\
    \subfloat[]
		{\includegraphics[width=0.45\columnwidth]{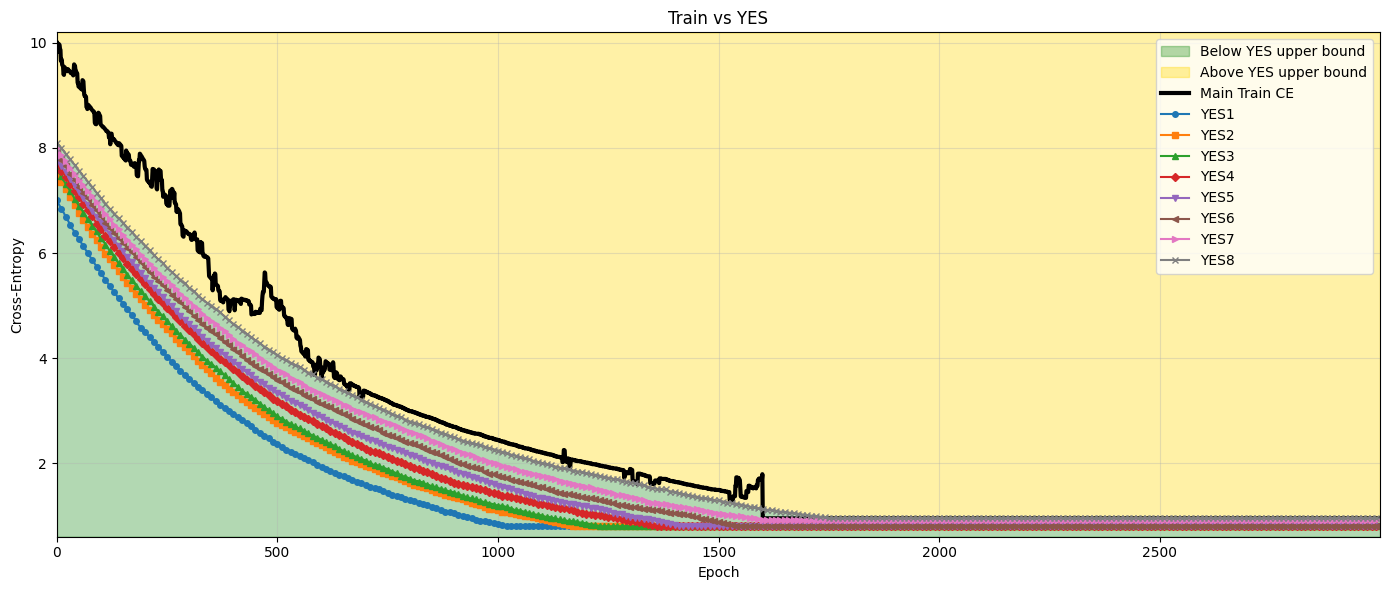}}\hspace{35pt}\quad
    \subfloat[]
		{\includegraphics[width=0.45\columnwidth]{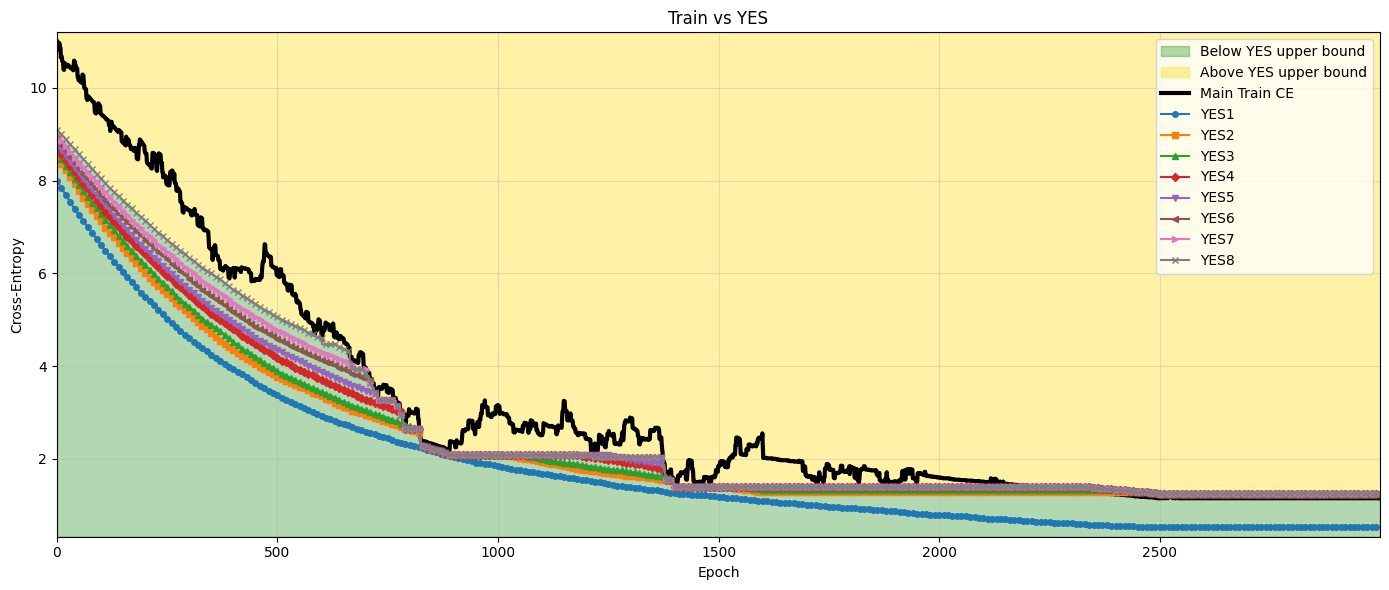}}
\caption{Fine-tuning the 26-layer OpenLLaMA model under quantization. Subplots (a,b) use ternary quantization with learning rates $2e-4$ and $1e-5$, respectively, while (c,d) use channel-wise binarization with the same learning rates. Training trajectories are compared against YES bounds: 5-layer bounds for (a,b) and 26-layer bounds for (c,d).}

\label{figure_3}
\end{figure*}

Fig.~\ref{figure_2} presents four subplots. (a) corresponds to a 4-layer LLaMA-style decoder with a learning rate of $2e-4$, while (b) shows a 4-layer GPT-2–style decoder with a learning rate of $1e-5$. For the YES baselines, we set $L' = 2$ for (a) and consider the following permutations:
YES1 = [1,2], YES2 = [2,2], YES3 = [1,3], YES4 = [1,4], YES5 = [3,3], YES6 = [4,4], YES7 = [2,4], and YES8 = [3,4].
For (b), the number of YES layers matches the main model, and only the permutations are varied: YES1 = [1,2,3,4] (direct permutation), YES2 = [3,3,3,4], YES3 = [4,4,4,4], and YES4 = [1,2,4,4].

For all YES configurations, the learning rate is set to $1e-3$ for hidden layers and $5e-4$ for the final linear projection to the vocabulary. The cache size equals the batch size, and each layer is optimized for six iterations.

In (a), the training loss decreases smoothly and surpasses all YES baselines except YES4 by epoch 15. This indicates that joint optimization across layers is more effective than most permutations, while the superior performance of YES4 = [1,4] highlights the dominant role of the first and last layers in this configuration. We emphasize that these results reflect training dynamics over a fixed number of epochs.

In (b), the training trajectory converges smoothly due to the chosen learning rate, while all YES baselines converge sharply and eventually coincide. This behavior suggests that, despite different permutations, the optimization reaches a stable solution, and according to our baselines, a well-performing model is obtained after 200 epochs.

For Fig.~\ref{figure_3}, we fine-tune the pretrained OpenLLaMA model with 26 layers on our dataset, applying quantization during training. Quantization is applied only to the weights of the MLP, gating layers, and the final linear projection. Fine-tuning is performed with an effective batch size of 64 sequences (512 tokens per sequence), implemented via gradient accumulation due to memory constraints. (a) and (b) use ternary quantization, while (c) and (d) use channel-wise binarization. The learning rates are $2e-4$ for (a) and (c), and $1e-5$ for (b) and (d). We employ a linear scheduler that decreases the learning rate by a factor of 0.98 every 50 epochs. The YES bounds use $L' = 5$ layers for (a) and (b), and $L' = 26$ for (c) and (d).  

For the 5-layer YES bounds in (a), the eight permutations are listed in Table~\ref{table_yes_compact}. We observe that ternary quantization yields smoother convergence than channel-wise binarization, as expected since ternary quantization can eliminate parameters that are redundant for this dataset. In subplot (a), the training trajectory struggles early and only surpasses all YES bounds except YES1 and YES2 after roughly 500 epochs; notably, YES1 closely tracks training until full convergence at 3000 epochs, indicating that the upper layers dominate solution quality in this setting. In subplot (b), the training curve lies below most YES results, except for YES2, YES5, and YES7. As training progresses, it gradually aligns with these bounds, with YES7 slightly outperforming the others and closely following the training trajectory. This behavior suggests that this particular permutation of layers captures critical aspects of the optimization dynamics.

\begin{table}[t]
\centering
\setlength{\tabcolsep}{2pt}
\renewcommand{\arraystretch}{0.9}
\caption{YES permutations used for Fig.~\ref{figure_3}. The notation $\times k \times n$ denotes that value $n$ is repeated $k$ consecutive times.}
\label{table_yes_compact}
\begin{tabular}{c|l|l}
\hline
\textbf{YES} & \textbf{10-layer} & \textbf{26-layer} \\
\hline
YES1 & [1,2,26,26,26] & [1,2,10,$\times$23×26] \\
YES2 & [2,3,6,8,10] & [11,12,13,14,15,16,17,18,$\times$7×25,$\times$11×26] \\
YES3 & [1,4,5,9,10] & [1,3,5,7,9,11,13,15,$\times$6×22,$\times$8×25,$\times$4×26] \\
YES4 & [3,4,6,25,26] & [1–10,$\times$5×17,$\times$3×18,19,$\times$7×26] \\
YES5 & [1,3,5,7,9] & [1,2,4,5,7,8,10,11,13,$\times$5×14,15,16,17,18,19,$\times$7×26] \\
YES6 & [2,4,6,25,26] & [3,4,6,7,9,10,12,13,$\times$6×24,$\times$9×25,$\times$3×26] \\
YES7 & [1,2,3,4,25] & [1,2,3,6,7,8,9,10,11,16,$\times$9×24,25,$\times$6×26] \\
YES8 & [10,10,25,25,26] & [1-26] \\
\hline
\end{tabular}
\vspace{-15pt}
\end{table}

For subplots (c) and (d), where the YES structure spans all 26 layers, training improves more slowly. Even after 600 epochs, it reaches only YES8, and by 1600–2300 epochs it approaches convergence without surpassing the stronger YES permutations. Interestingly, in subplot (d), all YES bounds closely track the training trajectory, including during temporary drops, with YES1 emerging as the strongest bound. Here, YES1 projects 23 Transformer layers to the output of the final layer, suggesting that most intermediate layers in this configuration are less influential and play a limited role in determining the final solution. Under quantization, representational bottlenecks become more pronounced. Later layers appear to absorb most of the expressive burden, while earlier layers contribute marginally. Importantly, a reader examining only the training curve in (d) might conclude that optimization has converged; however, the YES bounds reveal that better optimization paths may exist beyond the one selected by training.
\begin{figure}[t]
	\centering
	{\includegraphics[width=0.5\columnwidth]{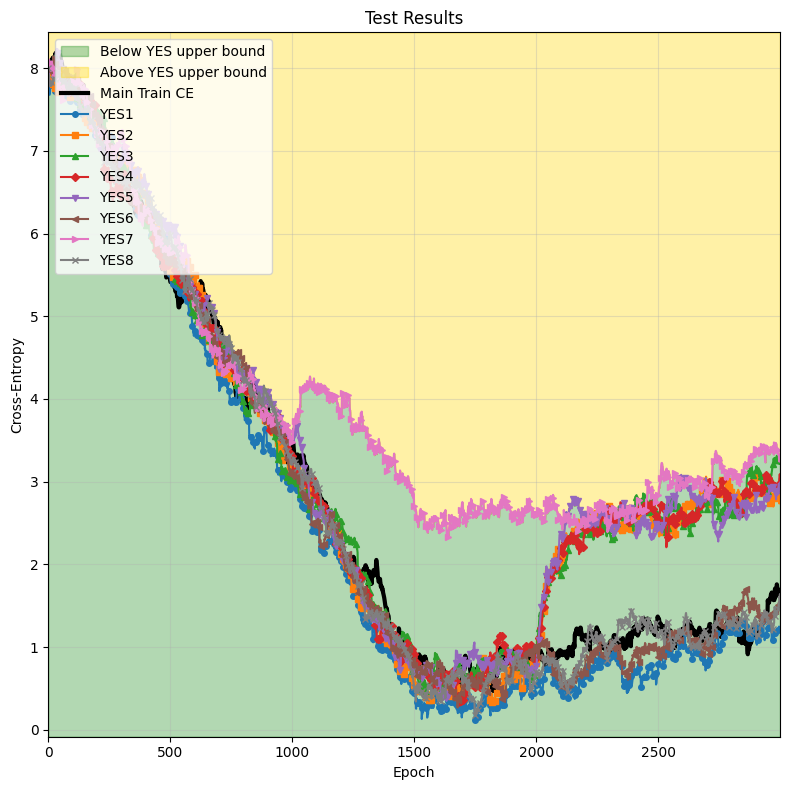}}
	\caption{Test results for train and YES solutions over 3000 epochs. The test results are obtained by evaluating each YES solution on the test dataset objective.}
\vspace{-10pt}
\label{figure_cloud}
\end{figure}

\subsection{Notable Observations}
In several subplots (notably Fig.~\ref{figure_3}(a) and Fig.~\ref{figure_3}(d)), the training curve continues to decrease for hundreds or thousands of epochs after it has already aligned with the strongest YES bound. This suggests that YES bounds can serve as an early stopping signal that is independent of validation loss. Once training meets the tightest YES bound, further optimization primarily refines within an already-certified region rather than discovering qualitatively better solutions. In Fig.~\ref{figure_3}(d), training exhibits visible drops and plateaus, yet the YES bounds track these dynamics tightly and remain informative throughout. This decouples optimization stability from solution quality. A smooth loss curve is not a reliable indicator of having reached a strong solution—especially under quantization. The fact that YES1 dominates in Fig.~\ref{figure_3}(d) implies that most intermediate layers are effectively redundant for this dataset and training regime. This is a data-driven argument for layer pruning or selective fine-tuning, derived without retraining multiple ablated models.

\subsection{Test Results Discussion}
For the test results, we observe some interesting patterns. In some cases, when certain YES solutions achieve better training performance than the train solution, their corresponding test results closely follow the train results, showing only minor differences. The test results are obtained by evaluating each YES solution using the objective function on the test dataset. However, when the train solution itself becomes better, its test performance can sometimes be slightly worse than that of some YES solutions, specifically those YES solutions that had superior training results. For example, Fig.~\ref{figure_cloud} shows the test results corresponding to Fig.~\ref{figure_3}(a). As seen, before approximately epoch 1500, the results behave very similarly, even though some YES solutions provide better training outcomes. After epoch 1500, some YES solutions perform worse than the train test results. Interestingly, YES1 consistently remains competitive on the test set and can even outperform the test results provided by the train solution of the model itself. Because the size of the dataset is small relative to the number of parameters in the model, overfitting would normally be expected. However, YES1 demonstrates better generalization on the test dataset compared to the train solution.

\bibliographystyle{unsrt}  
\bibliography{references}

\newpage
\appendix

\section{Toy Example: MNIST Dataset}

\begin{figure*}[t]
	\centering

    \subfloat[]
		{\includegraphics[width=0.68\columnwidth]{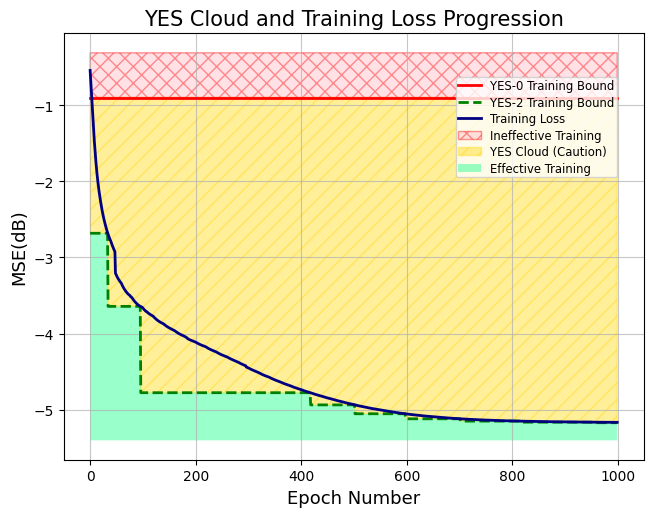}}\hspace{35pt}\quad
    \subfloat[]
		{\includegraphics[width=0.68\columnwidth]{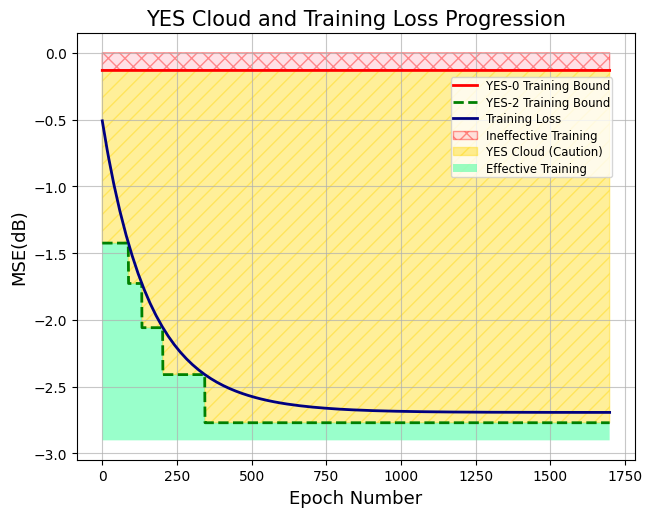}}
	\caption{Training trajectories of quantized FCNNs compared with the proposed YES bounds. Subplots (a) and (b) show a four-layer network trained with learning rates of $10^{-5}$ and $10^{-6}$, respectively. }
\label{figure_1}
\end{figure*}
\paragraph{Toy experiment: MNIST}
We begin with a simple toy experiment on the MNIST dataset to demonstrate the effectiveness of the proposed baseline in a low-capacity setting. In this experiment, we apply weight binarization during training and analyze its impact on convergence and stability. This controlled setup serves as a sanity check and highlights the robustness of the proposed approach under extreme quantization. 

For the toy classification experiment, we use the MNIST handwritten digits dataset, where grayscale images are normalized to the range $[0,1]$ and flattened into 784-dimensional vectors. We use a subset of 5{,}000 training samples, with labels encoded as one-hot vectors over 10 classes. The model is a fully connected feed-forward network with hidden layer widths of 100 and 50, followed by a 10-way output layer. Training is performed with a batch size of 1000, and this setup is used to analyze optimization behavior and training stability under weight binarization. Training is carried out under various parameter configurations using SGD. During pretraining, the learning rate is initialized as $\eta = \eta_0$ and decayed by a factor of $0.9$ every $50$ epochs. Additionally, channel-wise quantization is applied in the training setup. 

As training progresses, the YES bounds evolve alongside the training loss, providing a visual tool to monitor how optimization dynamics interact with certified baselines. This view highlights when the training trajectory crosses the YES-0 bound, surpasses tighter YES-$k$ bounds, and how these transitions occur over time.

The resulting cloud consists of the YES-0 bound and a sequence of progressively tighter YES-$k$ bounds. Its characteristic stair-step structure arises because bounds are updated only at epochs where the training trajectory first exceeds them. Figure~\ref{figure_1} illustrates this behavior for a four-layer network trained with learning rates of $1e-5$ and $1e-6$.

When the solver is well configured, the training trajectory converges to the same point as the YES bounds, certifying effective optimization behavior. In contrast, a smaller learning rate leads to slower or misaligned convergence, highlighting the sensitivity of stochastic gradient methods to learning-rate selection in quantized settings.

\end{document}